\documentclass[pdflatex,sn-mathphys-num]{sn-jnl}

\usepackage{graphicx}%
\usepackage{multirow}%
\usepackage{amsmath,amssymb,amsfonts}%
\usepackage{amsthm}%
\usepackage{lmodern}
\usepackage{fix-cm}
\usepackage[title]{appendix}%
\usepackage{xcolor}%
\usepackage{textcomp}%
\usepackage{manyfoot}%
\usepackage{booktabs}%
\usepackage{algorithm}%
\usepackage{algorithmic}%
\usepackage{listings}%
\usepackage{subcaption}%
\usepackage{array}%
\usepackage{booktabs}
\usepackage{longtable}

\begin{document}

\title[Fog of Love: Engineering Virtuous Agent Behavior with Affinity-based
  Reinforcement Learning in a Game Environment]{Fog of Love: Engineering Virtuous Agent Behavior with Affinity-based
  Reinforcement Learning in a Game Environment}


\author*[1]{\fnm{Ajay} \sur{Vishwanath}}\email{ajay.vishwanath@uia.no}

\author[1]{\fnm{Christian} \sur{Omlin}}\email{christian.omlin@gmail.com}

\affil[1]{\orgdiv{IKT}, \orgname{University of Agder}, \orgaddress{\street{Jon Lilletuns vei}, \city{Grimstad}, \postcode{4879}, \state{Agder}, \country{Norway}}}

\abstract{Instilling virtuous behavior in artificial intelligence has seen increasing interest. One of the techniques proposed is known as affinity-based reinforcement learning, which uses policy regularization on the objective function to incentivize virtuous actions without being fully dependent on the reward function design. Thus far, this technique has been demonstrated to be effective in grid worlds and toy-problem environments with minimal state and action spaces. To expand this research to more sophisticated environments, we introduce a two-player multi-agent environment based on the role-playing board game known as Fog of Love. In this environment, two agents compete to fulfill their individual virtues, while also cooperating to satisfy their relationship. Given the multi-agent nature, this is a complex problem where multi-agent deep deterministic policy gradient agents neither compete nor cooperate successfully. We present evidence that localized affinities enhance agent performance in achieving both competitive and cooperative objectives, resulting from superior overall scores in both domains. This not only results in virtuous choices but also clarifies an agent's teleology and makes its behavior human-level interpretable.}



\maketitle

\section{Introduction}\label{intro}

Programming artificial intelligence to make ethical decisions by embedding moral values and norms, falls under the umbrella of the discipline known as \textit{machine ethics} \cite{Anderson_Anderson_2007,wallach_moral_2010,zhong_computational_2025}. These moral values could be based on popular ethical theories such as deontology, consequentialist and virtue ethics, along with others such as moral particularism, prima-facie principles, etc. Whereas deontological ethics states that morally right or wrong actions are determined using rules and duties \cite{alexander_deontological_2021}, consequentialist ethics prioritizes the consequences of an action that causes the best outcome for the majority, which is in turn the morally praiseworthy act \cite{TH_sinnott-armstrong_consequentialism_2003}. Lastly, virtue ethics is a normative ethical theory positing that virtues constitute the pinnacle of moral character, whereby an individual exemplifying virtue actively endeavors to attain excellence in the demonstration of virtuous conduct. For instance, a brave person would know the amount of bravery to exhibit based on moral exemplars or role models who are brave, and from their own individual experience. This balance of extremes is referred to as the \textit{golden mean}.

Ancient philosophers such as Aristotle \cite{ross_oxford_1980} have given a comprehensive analysis of virtues (e.g., temperance, honesty and bravery), leading to the flourishing (\textit{eudaimonia}) of virtuous individuals in Athenian society. Today's virtue theories build upon the Aristotelian framework; they emphasize different virtues and may include socio-cultural nuances \cite{hursthouse_virtue_2001,crisp1997virtue}. In the ethics of technology discipline, arguments for the plausibility of virtuous behavior in artificial intelligence (AI) have been made, especially because of a virtue ethics feature known as \textit{phronesis} \cite{stenseke_artificial_2021,vishwanath_towards_2022}. A young, inexperienced person seeks role models to strive towards virtuous conduct. They develop their character with more life experience, coupled with inspiration from their role models, such as their parents, teachers, leaders or highly acclaimed historical figures. This process of reflection and self-improvement is known as \textit{phronesis}.

Reinforcement learning (RL) has been proposed as technique to emulate \textit{phronesis} due to its ability to learn from a set of states, actions and rewards, to compute an optimum strategy \cite{stenseke_artificial_2021,vishwanath_towards_2022,guarda_machine_2024}. However, in its most basic form, RL is purely utilitarian, which relies entirely on its reward function. To reduce this reliance,  affinity-based RL (ab-RL) \cite{vishwanath_exploring_2024} was recently proposed, which uses a prior probability distribution of actions to regularize an agent's objective function. In the context of virtuous agents, the prior probabilities act as role models to an RL agent. An agent interacts with its environment, to find a balance between its priors and its reward function, similar to how a virtuous person might find the \textit{golden mean} of extremes; these extremes are considered \textit{vices} in the virtue ethics paradigm. These action priors could depend on the scenario, such that agents \textit{prefer} some actions over others, because it is an act a role-model would have committed in a similar situation.

Role-playing computer games such as \textit{The Witcher 3: Wild Hunt} \cite{cd_projekt_red_witcher_2015}, Disco Elysium \cite{zaum_disco_2019}, \textit{This War of Mine} \cite{11_bit_studios_this_2014}, and board games such as \textit{Fog of Love} \cite{jacob_jaskov_fog_2017}, and \textit{Dead of Winter} \cite{jonathan_gilmour_dead_2014}, have become popular. In these games, players need to resolve moral dilemmas which affect the future states and actions. Some games such as \textit{Life is Strange} and \textit{Mass Effect} compel players to reflect further on their virtues, whose outcomes are independent of the choices \cite{nay_meaning_2017}. There are other games such as \textit{Papers, Please} where the moral dilemmas may not be immediately apparent while playing the game but will manifest at a later stage of the game, known as \textit{systemic} role-playing games \cite{formosa_papers_2016}. These mechanisms within such games make them popular and entertaining, while also nudging ethical behavior. With the success of deep reinforcement learning solving Atari games, chess and Go, makes the exploration of role-playing games with moral dilemmas an interesting proposition \cite{mnih2013playing,silver2018general}. 

In this paper, we leverage the game mechanics and dynamics of Fog of Love \cite{jacob_jaskov_fog_2017}, a two-player role-playing board game with competitive and cooperative goals. The players are in a relationship and they navigate scenarios in the game such as choosing between their personal goals (competitive) and their partner's needs (cooperative). The competitive goals are random and a player strives to reach their trait goals such as \textit{discipline}, \textit{sensitivity} and \textit{gentleness}. Simultaneously, the players make decisions which are aligned with their partner's choices thus increasing their \textit{satisfaction} score. We develop an RL environment based on the core mechanics in Fog of Love. 

The rest of the manuscript is organized as follows: Section \ref{relworks} situates our work in the larger context of machine ethics and reinforcement learning. Section \ref{fol} describes the technical aspects of Fog of Love, where we outline the game mechanics, introduce the RL environment and define evaluation metrics. Section \ref{algos} presents details of the affinity-based RL paradigm which uses multi-agent deep deterministic policy gradient algorithm. A comparison of our localized (state-dependent) ab-RL algorithm with a baseline algorithm in Section \ref{results} demonstrates that virtuous behavior can be imprinted without tedious reward engineering. Our results show a significant improvement in performance, in regularized competitive goals and cooperative goals. In Section \ref{discussion} we reflect on the the significance, impact and implications for machine ethics, in light of our results. 

\section{Related Works} \label{relworks}

In this section, we summarize research related to machine ethics and reinforcement learning.

\subsection{Machine ethics and artificial virtuous agents}

Machine ethics has been an active research field since its introduction in 2008 through a book by Wallach and Allen \cite{wallach_moral_2010} titled ``Moral Machines: Teaching robots right from wrong''. Since then, several other researchers, such as Anderson \& Anderson \cite{Anderson_Anderson_2007}, Dennis et al \cite{DENNIS20161}, Moor et al \cite{moor_nature_2006}, made significant contributions to this field. According to landmark surveys \cite{cervantes_artificial_2020,tolmeijer_implementations_2021,zhong_computational_2025} most research in this field used deontological \cite{peschl_moral_2022} and consequentialist ethical theories \cite{rodriguez-soto_instilling_2022} to develop morally salient agents. For instance, Ozaki et al \cite{ozaki_finding_2024} develop an algorithm which calculates the middle ground between deontological norms designed to navigate through moral dilemmas. They empirically verify their work on the Moral Machine Experiment \cite{awad_moral_2018}, which is a platform that gathers decisions on moral dilemmas faced by autonomous vehicles. Others \cite{lang_utilitarian_2002} have applied a utilitarian model to perform non-monotonic reasoning on desires and goals. There are several other machine ethics contributions in these ethical theories and has been pointed out that this is mainly because these theories can be formalized and codified \cite{stenseke_artificial_2021}. 

There have been  very few implementations of virtue ethics because the description of virtues is \textit{vague}; it is a subjective trait which involves feelings, reflection, learning, and cognition, which are often challenging to measure, let alone, engineer. However, there have been explorations in the development of virtuous agents, whose definition is based on \textit{functional morality} by Wallach and Allen \cite{wallach_moral_2010}. Functional morality entails a significant degree of an artificial agent's autonomy and moral sensitivity, although it does not exhibit the full extent characteristic of human moral agents. Some of the research based on Wallach and Allen's definition of morality proposes virtue frameworks and architectures \cite{guarda_machine_2024,govindarajulu_toward_2019}, while others are empirical evaluations of virtuous behavior\cite{stenseke_artificial_2022,vishwanath_exploring_2024}.

Stenseke \cite{stenseke_artificial_2021} proposes to implement an artificial virtuous agent based on a \textit{connectionist} approach, which dedicates a multi-layer perceptron for each virtue. For example, separate networks for honesty and courage are trained if the environment contains elements related to honesty and courage, respectively. Together, these virtues form a ``virtue'' network, which informs the agent's action. The performed action is evaluated by a value function, which in turn informs a \textit{phronetic} system which updates the virtue network. This was evaluated in Tragedy of the Commons based environment called \textit{BridgeWorld} where agents either consume or share limited resources. The results in comparison to the baseline proved that selfish-selfless hybrid agents performed the best, with minimum \textit{death rate} \cite{stenseke_artificial_2022}.

An implementation of artificial virtuous agents by Vishwanath and Omlin \cite{vishwanath_exploring_2024} demonstrated the use of affinity-based reinforcement learning which uses policy regularization of the action space. They developed an environment based on the role-playing game \textit{Papers, Please}, where the player is an immigration officer must process entrants into their country. Every time the entrant is processed correctly, they are paid a salary, and if they process incorrectly, they are penalized. Based on their daily earnings, the player must decide whether to feed their family or provide heat. Sometimes while processing entrants, the player might be offered a bribe, which makes it easier to earn money and feed their family. The authors used affinity-based reinforcement learning to guide the player towards honest and compassionate actions such as to refuse bribes and feed the family whenever possible while also saving their salary \cite{vishwanath_exploring_2024}. However, only one moral dilemma was resolved, and it lacked complexity of multiple dilemmas, and had a limited action space.

Another approach to engineer virtuous behavior proposes inverse RL \cite{berberich_virtuous_2018} based on Aristotle's virtues such as prudence, honesty, friendship to humans, and temperance. While inverse RL captures human demonstrations, it is challenging to account for every possible scenario, and, given the sophisticated nature of inverse RL algorithms convergence could be a challenge \cite{adams_survey_2022}. An alternative was proposed in \cite{govindarajulu_toward_2019} to develop a formal model of virtuous behavior based on virtues and vices. While these are good first steps, these contributions require empirical evaluations either through simulated environments or in the real world. In our work, we develop a game environment where competing and cooperating agents face moral dilemmas that they need to resolve.

\subsection{Reinforcement learning and desired behavior}

In the domain of machine ethics, Rodriguez et al. \cite{rodriguez-soto_instilling_2022} theorized and implemented a multi-objective RL (MORL) algorithm with a reward objective and an ethical objective. They demonstrated its performance in a Public Civility environment, where the primary objective is to move from point A to point B, and at the same time, the agent could dispose trash on the way. Using MORL, the researchers achieved both objectives. However, this is a purely utilitarian approach that involves defining a second reward function to achieve the ethical goals. We will see below why this approach utilizing only reward functions could be an issue.

Training RL agents to exhibit behaviors presents significant challenges. This difficulty is particularly pronounced due to the RL agents' dependence on a reward function, which frequently acts as a barrier to achieving the specified behavioral objectives \cite{ng1999policy}. Reward shaping is often used as a method to steer an agent towards desired behavior by using intermediate goals. A challenge with this technique, however, is that an agent might end up exploiting the intermediate rewards rather than the primary goals. Furthermore, reward shaping exacerbates the lack of explainability of agent behavior \cite{reward_hacking}. Another approach known as preference-based RL utilizes preferences instead of numerical reward functions \cite{wirth2017survey}. This way, an agent's behavior can be elegantly explained because its policy is dictated by preferences demonstrated by a human expert. However, an issue with preference-based RL is its reliance on laborious and detailed demonstration by human experts, sample inefficiency and the computational complexity of learning from preferences instead of numeric rewards. Another challenge is its performance in continuous state and action spaces.

As discussed, reward shaping and preference-based RL  \textit{encourage} agents towards their goals through differing means. An alternate approach is to \textit{prevent} catastrophic state-action combinations through safe RL and constrained RL \cite{garcia2015comprehensive,achiam2017constrained}, which avoid undesirable states or actions using varying ways in their environment. While these techniques may employ reward-shaping and use safety layers to configure risk-aware policies, they can be restrictive to exploration of the environment. In techniques such as policy regularization, an agent is encouraged to perform exploration. Regularization techniques can impart advantageous characteristics to RL agents by incorporating prior information to inform policy optimization, which is particularly advantageous in multi-task and transfer learning scenarios \cite{ran2023policy}. This approach mitigates the issue of value overestimation prevalent in offline RL settings and enhances policy explainability without necessitating modifications to the underlying learning algorithms \cite{persiani2022policy}. Furthermore, policy regularization facilitates stable control in continuous action spaces by imposing penalties on excessive deviations from optimal action trajectories \cite{tirumala_behavior_2022}.

In this paper, we adopt the policy regularization philosophy in the form of affinity-based RL (ab-RL). This technique was earlier utilized in grid-world environments \cite{maree_reinforcement_2022}, toy problems \cite{vishwanath_exploring_2024}, and data-driven financial investment recommendation \cite{maree_can_2022}. In these works, ab-RL successfully achieved its objectives while also demonstrating affinities towards certain combination of action probabilities. For example, in the financial investment problem \cite{maree_can_2022}, depending on the age, risk profile and personality, ab-RL was used to achieve high return on investments while also being cognizant to the other factors. In a similar vein, researchers \cite{vishwanath_exploring_2024} experimented in a simpler environment but riddled with moral dilemmas. They demonstrated that virtuous behaviour was plausible using ab-RL and that varying configurations of the same could be achieved using hyperparameter tuning. An update to ab-RL was proposed \cite{vishwanath_localized_2025} that incorporated state-dependent context. These regularization-based approaches differ from constrained optimization because rather than being reward-based constraints, the agent is predisposed with affinities that allow it to explore the environment and determine an optimal action. In this manuscript, we define a complex environment, such that a different set of virtues need to be fulfilled in different episodes. Hence the agent learns to anticipate and model these changing dynamics. Below we outline the Fog of Love environment.

\section{The game: Fog of Love} \label{fol}

In this section, we describe the Fog of Love game mechanics, dynamics, the reinforcement learning environment and introduce evaluation metrics.

\subsection{Game mechanics}

\begin{figure}
    \centering
    \includegraphics[width=0.7\linewidth]{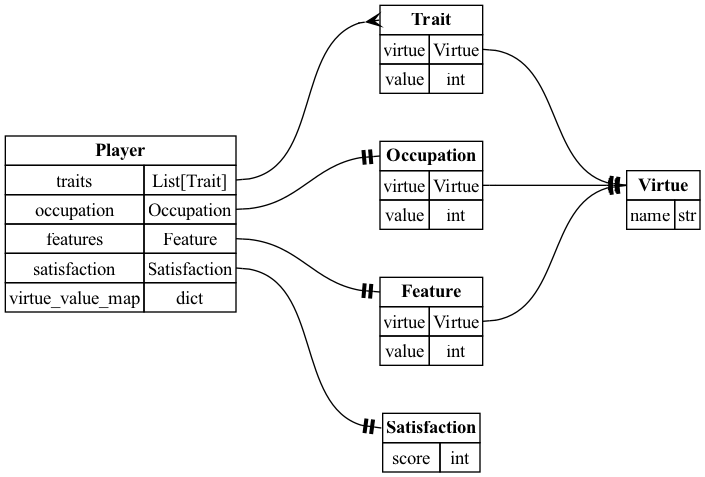}
    \caption{\textbf{Entity relationship diagram specifying the relationship between a player, traits, occupation, feature, satisfaction and virtues:} A player can have one and only one occupation, feature and satisfaction score. While they can have several trait goals (upto six). The traits, occupations and features can have any of the six virtues with an integer value. The virtue\_value\_map contains a dictionary which tracks the current value of the trait the player has earned in the game.}
    \label{fig:player}
\end{figure}

Fog of Love is a two-player game with cooperative and competitive goals to fulfill. This will be a technical description of the game mechanics and while it does not cover every aspect of the game, it discusses the core aspects which largely influence Fog of Love.  First, Player 1 randomly chooses an occupation and three \textit{Trait} goals. This is followed by Player 2 choosing a \textit{feature} they liked about the Player 1 out of three random feature cards. Each occupation, trait and feature card contains a virtue such as discipline, sincerity, sensitivity, extraversion, gentleness, and curiosity, and an integer value associated with said virtue. For example, the player could pick out three \textit{Trait} cards with $discipline=4$, $extraversion=-5$ and $gentleness=2$, which is something the player must achieve at the end of the game. In Figure \ref{fig:player}, we highlight these components relative to the player. The $virtue\_value\_map$ is initialized with a virtue level based on the player's occupation and feature. And this map is updated throughout the game, and used to determine whether a player has accomplished their \textit{Trait} goal. And finally, \textit{satisfaction} is a variable that keeps track of how much a player is satisfied with the relationship. Player 2 repeats the same process as Player 1.

\begin{figure}
    \centering
    \includegraphics[width=0.8\linewidth]{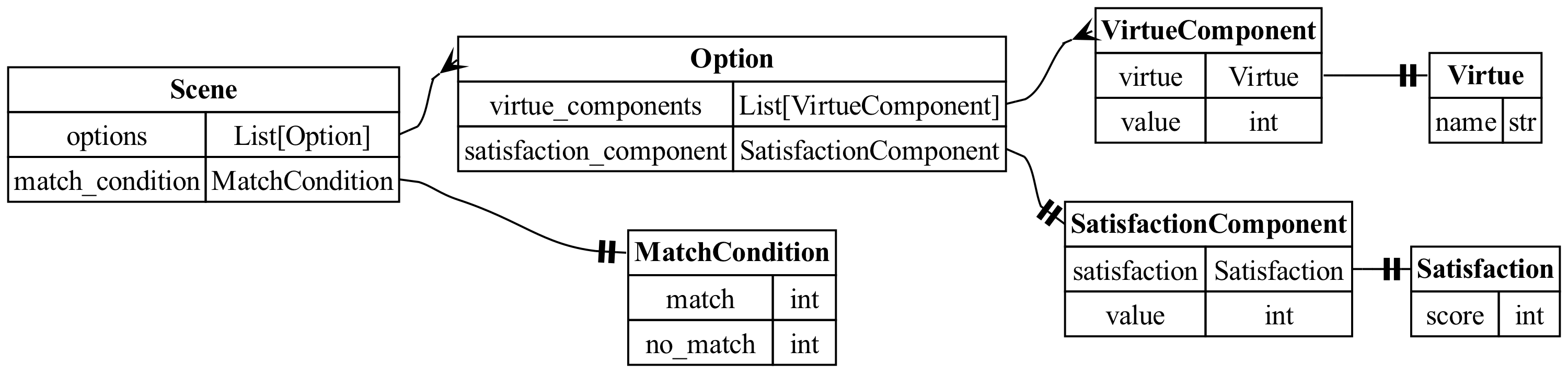}
    \caption{\textbf{Entity relationship diagram of a \textit{scene} with \textit{options} and \textit{match conditions}}: A scene consists of multiple options. Each option can have several virtue components and one satisfaction component. Also a scene has a match condition which specifies the reward and penalty if the player actions are the same and if they differ, respectively. }
    \label{fig:scene}
\end{figure}

After the players know their trait goals and their $virtue\_value\_map$ is initialized, a \textbf{Scene} card is played, where each scene describes a scenario or a question. Here, one or both players choose an answer at the same time from the options (A, B, C, etc). Each option can contain up to three virtue-value pairs, and if a player chooses this option, their own $virtue\_value\_map$ is updated with the virtue component they chose.  An option choice also affect the  player's \textit{satisfaction} value, which then will then determine how much the player prioritizes cooperation within the relationship. Finally, the \textit{match condition} is an additional component of a \textit{scene} where, usually, when the players' choices align, then they are rewarded with positive \textit{satisfaction}, or negative reward if their choices do not align. In Figure \ref{fig:scene} we highlight these relationships. 

A \textit{chapter} can contain upto 10 \textit{scenes} where at the end of every chapter the \textit{Scene} cards are discarded and new ones relevant to the new chapter are drawn. The game ends when all chapters are played, and if the players fulfill their \textit{destiny}. A destiny can be along the lines of ``both players fulfill $satisfaction=30$'', ``Player 1 fulfills $extraversion=5$'', and ``Player 2 fulfills $sensitivity=-7$'', for example. A player can choose to draw, keep and discard destiny cards throughout the game. If both players fulfilled their \textit{destiny}, then the relationship survives. The main challenge in this game is that the players are not aware of each other's \textit{trait} goals; hence a player must choose actions which should not sabotage the opponent's trait goals while accomplishing their own \textit{trait} goals. At the same time, they must also improve their \textit{satisfaction} scores, which increases when they select the same \textit{option} in a \textit{scene}.

\subsection{Reinforcement learning environment}

We implemented a Fog of Love environment using OpenAI gym framework\footnote{Available at https://github.com/ajvish91/fog-of-love-rl}. Below, we define the state space, action space, reward function, environment dynamics, and evaluation metric. We conclude this topic by discussing the elements in the game that were \textit{not} included.

\begin{itemize}
    \item \textbf{Observation space}: The observation space contains 88 observations, 44 for each player. For each player, their $virtue\_value\_map$ consisting of six virtues each is known to both players, hence making it 12 observation variables per player. The same applies for each player's \textit{satisfaction} score. Next, their respective \textit{trait} goals and \textit{satisfaction} goals are included in the observation space which are constant throughout the game. The agents are not privy to the other's \textit{trait} or \textit{satisfaction} goal, however, both agents need to fulfill their \textit{satisfaction} goal every episode. Finally, a \textit{scene} is part of the observation space which consists of three \textit{options} and each option contains a value of every virtue and one \textit{satisfaction} value. In total, the three \textit{options} combine together to form 21 observation variable per player. Finally, a match condition is included where two observation variables specify the value rewarded or penalized when their \textit{option} choices are the same or different, respectively. We present details on the observation space in Appendix \ref{state-space}.
    \item \textbf{Action space}: Each agent has to choose between three actions which correspond to the respective \textit{option} generated within a \textit{scene}. Thus, the size of the action space is 3 for each agent and when a player chooses an action, the choice is deterministic.
    \item \textbf{Reward function}: The reward function is designed to incentivize the agent to achieve their \textit{trait} and \textit{satisfaction} goals. The agent is rewarded positively, if their \textit{option} choice aligns with their goal, and negatively if they choose the opposite. For example, if the agent's \textit{trait} goal is to achieve a \textit{gentleness} of $+7$, and if the agent chooses the option which contains $gentleness=+1$, then the agent is rewarded $+1$. On the contrary, if the option contained $gentleness=-1$, then the agent would be penalized by $1$. 
    \item \textbf{Environment dynamics: } Each episode in the game contains 100 \textit{scenes} or turns, where both agents choose their \textit{option} at the same time. The observation space is updated based on the choices made by the agents. Also of significance is that the \textit{scenes} are generated at random, with random virtues and satisfaction values. Hence, this is a highly stochastic environment, and is challenging for an agent to \textit{hack} the reward function or the action space.
\end{itemize}


We exclude the notion of \textit{chapter}, \textit{destiny} and the dramatic elements in Fog of Love. However, we do loosely represent the \textit{destiny} component by setting the satisfaction goal as 30 for both agents, which is similar to the "Equal Partners" destiny in the original game. Overall, we believe that this RL environment does justice to the core components of the game, and the cooperative and competitive elements of the game.

\subsubsection{Example game-play}

Fog of Love is a game where both players are presented with a situation simultaneously. The game starts with random initial traits and trait goals assigned to both players (see Figure \ref{fig:player}). For example:

\begin{itemize}
    \item Player 1 occupation: \textit{gentleness} = -2
    \item Player 1 feature: \textit{sensitivity} = -1
    \item Player 1 trait goals: \textit{sincerity} = +5, \textit{gentleness} = -3
    \item Player 2 occupation: \textit{discipline} = +1
    \item Player 2 feature: \textit{discipline} = +1
    \item Player 2 trait goals: \textit{curiosity} = +4, \textit{sincerity} = -4
    \item Player 1 and Player 2 \textit{satisfaction} goal: +30
\end{itemize}

The \textit{satisfaction} values for both players are initialized to 0. Player 1 starts with \textit{gentleness} = -2 and \textit{sensitivity} = -1, while Player 2 starts with \textit{discipline} = +2, based on their occupation and feature values. The remaining virtues are initialized to 0. Supposing in the first round, the players receive the following three options:

\begin{enumerate}
    \item \textit{curiosity} = +1
    \item \textit{satisfaction} = -1
    \item \textit{discipline} = +1 and \textit{gentleness} = -1
\end{enumerate}

Also, if both players choose the same option, they are rewarded +2 satisfaction, and -3 otherwise. If Player 1 chooses Option 3 and Player 2 chooses Option 1, their respective virtues get updated with the value in that option. In other words, Player 1 gains +1 \textit{discipline} and -1 \textit{gentleness}, which is a good choice since Player 1 needs to fulfill \textit{gentleness} = -3. On the other hand, Player 2 chooses Option 1 which increases their \textit{curiosity} by +1, also good choice, given their trait goals. Both players receive a reward of +1 based on these choices\footnote{Since Player 1 chose Option 3, the fact that \textit{discipline} is not part of their trait goals does not affect the reward. However, were it the case that Option 3 contained \textit{sincerity} = -1 instead of \textit{discipline}, then the net reward for choosing Option 3 would be 0.}. However, since both players chose different options, their \textit{satisfaction} suffers a penalty of -3, which in the long term is not helpful towards achieving their \textit{satisfaction} goals. This sequence is repeated 100 times and the players are evaluated. If they achieve all their goals, then they have won the game and their `relationship' is a success.

\subsection{Evaluation Metrics}
A fundamental approach to assess this environment involves calculating the rewards obtained by the agents. However, this metric does not provide a comprehensive evaluation, as there are additional variables that can be used to ascertain whether an agent has successfully completed the game. A better method for this assessment is to evaluate whether the agent has achieved its predefined \textit{trait} and \textit{satisfaction} objectives. The \textit{success rate}, denoted as $\varrho$, is computed as the ratio of the number of \textit{trait} and \textit{satisfaction} goals attained to the total number of goals established within a given episode, formalized as follows:
$$\varrho = \frac{\text{goals achieved}}{\text{total number of goals}}$$
Although $\varrho$ provides a satisfactory high-level assessment of an agent's performance in the game, it lacks the granularity required to differentiate between agents with comparable overall success rates. To address this limitation, we introduce a metric termed goal error, $\delta_{err}$, which measures the extent to which an agent deviated from its intended objectives. We define goal error as follows:
$$ \delta_{err} = \sum_{i=1}^{N} \left| v_i - v_i^* \right| 
$$
Where $N$ is the number of goals that were \textit{not} achieved, $v_i$ is the actual value and $v_i^*$ is the target value of  $i$th goal which can be either \textit{virtue} and \textit{satisfaction} goals. For example, if a player's goal is $gentleness=4$, $satisfaction=20$ and $sensitivity=-5$, and if they got scores of $gentleness=5$, $satisfaction=4$ and $sensitivity=2$, they only succeeded in achieving $gentleness$. This means, $\varrho$ will be 0.33 and $\delta_{err}$ is $\left| 20-4 \right| + \left| -5 - 2 \right|$ which sums up to 23. We compute the mean values across episodes to evaluate the performance of different agents with varying affinities.

\section{Algorithms} \label{algos}

In this section, we formally describe the RL algorithms we implement and specify how they were modified to play in the Fog of Love environment. First, we describe multi-agent deep deterministic policy gradient (MADDPG) algorithm, which we use as the baseline for comparison. Next we define policy regularization and affinity-based RL. Finally, we define and outline localized affinity-based RL.

\subsection{Multi-agent deep deterministic policy gradient (MADDPG)}
Based on the single-agent policy gradient algorithm, specifically the deep deterministic policy gradient (DDPG) algorithm, we employ the multi-agent variant referred to as MADDPG. DDPG operates using an actor-critic architecture, wherein the actor processes environmental observations to output the probability distribution over a discrete set of actions (policy), while the critic evaluates the action's quality through the computation of its associated Q-value. The following section delineates the gradient of the objective function utilized by DDPG:
\begin{equation} \label{eqn:ddpg}
\nabla_{\theta} J(\theta) = \mathbb{E}_{s \sim D} \left[ \nabla_{\theta} \mu_{\theta}(a|s) \nabla_{a} Q^{\mu_{\theta}}(s,a) |_{a=\mu_{\theta}(s)} \right]
\end{equation}
where  $\mu_{\theta}(a|s)$ is the policy from which actions are selected, $Q^{\mu_{\theta}}(s,a)$, is the state-action value function, and $\mathbb{E}_{s \sim D}$ is the estimate of their product based on the observation sampled from a replay buffer which stores the state transitions. Here, policy $\mu$ and value function $Q$ are derived from the actor and critic respectively. In a multi-agent setting, each agent has its own actor while sharing a common critic \cite{lowe2020multiagentactorcriticmixedcooperativecompetitive}. Equation \ref{eqn:ddpg} becomes:
\begin{equation}\label{eqn:maddpg}
    \nabla J(\theta) = \mathbb{E}_{o,a \sim D} \left[ \nabla_{\theta_i} \mu_{\theta_i}(a_i|o_i) \nabla_{a_i} Q^{\mu_{\theta_i}}(o_i, a_1, \ldots, a_N) |_{a_i = \mu_{\theta_i}(o_i)} \right]
\end{equation}
where $o_i$ denotes the observational input specific to the $i$th agent, amidst a total of $N-1$ other agents. It is important to note that the Q-value function incorporates actions from all agents concurrently.

\subsection{Affinity-based RL}
Maree et. al \cite{maree_reinforcement_2022} introduced affinity-based RL (ab-RL) by incorporating a regularization term, $L$ to the objective function of an RL algorithm with a reward function $R(S, A)$, defined below:
\begin{equation}\label{eqn:affinity_obj}
    J(\theta) = \mathbb{E}_{S,A \sim D} [R(S, A)] - \lambda L
\end{equation}
\begin{equation}\label{eqn:affinity}
    L = \frac{1}{M} \sum_{j=0}^M \left[ \mathbb{E}_{A \sim \pi_{\theta}} [a_j] - (a_j | \pi_0(A)) \right]
\end{equation}
where $\pi_0(a)$ denotes the prior probability distribution of actions, and $L$ is computed by calculating the difference between the sampled estimate of $a_j$ (the $j$-th action among $M$ possible actions) derived from the policy and the prior probability of $a_j$, aggregated across all $M$ actions. A regularization hyperparameter $\lambda$ in Equation \ref{eqn:affinity_obj} is incorporated to modulate the intensity of the regularization effect. This policy regularization results in agents having an \textit{affinity} towards certain actions, thus making it possible to incentivize agents to behave a certain way without dependence on the reward function $R(s,a)$, while also being interpretable. 

Vishwanath et al \cite{vishwanath_localized_2025} expanded the regularization term to include local context, i.e., by performing state-dependent regularization. Hence, Equation \ref{eqn:affinity} can be rewritten as:

\begin{equation} \label{eqn:state_affinity}
    L_s = \frac{1}{M} \sum_{j=0}^{M} [E_{S, A \sim \pi_\theta}[a_j] - (a_j|\pi_{0i}(S, A))]^2
\end{equation}
where both the state, $S$ and action, $A$ are estimated from the policy $\pi_\theta$, and prior probability distribution $\pi_{0i}$ is a function of both $S$ and $A$. $L_s$ is the mean-squared error across all actions and states. Empirical evaluation on the Manhattan pizza delivery environment \cite{vishwanath_localized_2025} shows that an agent can be made to have varying affinities in different regions within the grid. In Fog of Love, we use an affinity condition to determine $a_j|\pi_{0i}(S, A)$. 

\begin{algorithm} 
\caption{Check affinity condition}
\begin{algorithmic}[1]
\label{alg:condition}
\REQUIRE state: list of values corresponding to the observation space, affinity\_idx: index representing the goal, option\_indices: indices of the options, current\_virtue\_index: index of the virtue in question.
\ENSURE tuple containing the option index (1, 2, or 3) and a boolean indicating if the condition is met
\STATE goal $\leftarrow$ \text{state[affinity\_idx]}
\STATE options $\leftarrow$ \text{state[option\_indices]}
\IF{goal $=$ 0}
    \RETURN $(0, \text{False})$
\ENDIF
\STATE goal\_positive $\leftarrow$ \text{goal $>$ 0}
\STATE virtue\_value $\leftarrow$ \text{state[current\_virtue\_index]}
\IF{(goal\_positive \AND goal $>$ state\_value) \OR (\NOT goal\_positive \AND goal $<$ virtue\_value)} 
    \FOR{$i, option$ \textbf{in} \text{enumerate(options, start=1)}}
        \IF{(goal\_positive \AND option $>$ 0) \OR (\NOT goal\_positive \AND option $<$ 0)}
            \RETURN $(i, \text{True})$  \COMMENT{(i, True) means option $i$ (1 ,2 or 3) meets the condition}
        \ENDIF
    \ENDFOR
    \RETURN $(0, \text{False})$ \COMMENT{(0, False) means no option meets the condition}
\ENDIF
\RETURN $(0, \text{False})$
\end{algorithmic}
\end{algorithm}

We define a novel function (Algorithm \ref{alg:condition}) which consists of the condition used to determine the affinities depending on the state and the action. When an option favours the affinity goal, then an affinity loss is incurred if the actor's current policy does not allocate a high probability of choosing this option. For example, if the agent receives three options (Option 1: \textit{sensitivity} = -1, Option 2:  \textit{gentleness} = +1, satisfaction = -1, and Option 3: \textit{satisfaction} = +1) and the agent's goal is to achieve \textit{gentleness} = +5, then the agent is made to favor Option 2 using Algorithm \ref{alg:condition}. The affinities are defined in terms of probabilities (for example [0.2, 0.6, 0.2]), such that Option 2 is preferred with a probability of 0.6, while the other options are to be chosen with a probability of 0.2 based on Equation \ref{eqn:state_affinity}. If the probability distribution of the sampled policy is (0.32, 0.05, 0.63), then the objective function sees a high affinity loss in addition to the actor and critic losses. In the next section, we show some results, comparing a baseline MADDPG with localized affinity-based MADDPG.

\section{Manifestation of Virtues} \label{virtues}

We defined the Fog of Love environment in Section \ref{fol} and affinity-based RL environment in \ref{algos}. Here, we make a case for the development of artificial virtuous agents using the game environment and the RL algorithm. While we do not aim to present a comprehensive philosophical treatise on virtue ethics, we ground our work on a specific moral theory and briefly provide reasons for why our implementation resonates with the same.

Aristotle based his definition of virtues on Athenian aristocrats \cite{ross_oxford_1980} while not applying the same to women and slaves. He argued for a virtue ethical framework based on the reality in ancient Greece by defining virtues such as honesty, temperance, friendship, and practical wisdom. MacIntyre \cite{macintyre_after_2007}, a contemporary philosopher, argued that virtues exist within social practices--a cooperative activity between humans with rules, individual and shared objectives. He compared the Aristotelian definition with those of Homer, Jane Austen, and Benjamin Franklin, and concluded that the nature of virtues can vary between social practices within a moral tradition. A person belonging to a specific moral tradition must reflect on their participation in social practices, in terms of a narrative, and based on whether this involvement allows them to fulfill their virtues.

MacIntyre presented an example from chess, where virtues such as ``analytic skill, strategic imagination, and competitive
intensity'' \cite[p.188]{macintyre_after_2007} are internal to chess. For example, a person involved in the social practice of chess could also involved in others such as being a healthcare practitioner (compassion, sincerity, honesty), and a snowboarder (flexibility, ambition, bravery). Virtues internal to each of these practices must be reflected upon in terms of a narrative: does my partaking in these practices help me improve my virtues? While one might argue that MacIntyre's virtue theory might be extremely relativist because we are divorcing ourselves from a universal moral theory, this is a misunderstanding because MacIntyre provides an objective moral framework within social practices to live a good life and strive towards a \textit{telos} within and across social practices.

In a similar vein, AI development can be argued to exist within MacIntyre's virtue framework. While we acknowledged in Section \ref{relworks} that humans and AI are not the same entities, rather AI can exhibit functional morality \cite{wallach_moral_2010}, based on what it perceives and how it acts. This allows us to consider AI algorithms within social practices, that could theoretically excel at virtues internal to a given practice while also achieving its objectives. However, AI, in its current form, cannot reflect on where it is being applied or whether its virtues are being utilized properly. Nevertheless, the onus is on the makers and engineers of the AI model who could contemplate the same in the environment within which the algorithm is deployed.

We argue that the virtues that exist as numbers within Fog of Love is compliant with MacIntyre's definition of virtue ethics. Using affinity-based RL we predispose agents to excel at \textit{sensitivity}, \textit{gentleness} or \textit{satisfaction}. While being good at achieving their trait goals (\textit{telos}), the ab-RL agent can also be trained to be cooperative with the other agent. In the next section, we present empirical evidence that our agents indeed can be trained to be artificial virtuous agents.

\section{Results}\label{results}

In this section, we present the results of the algorithms discussed here on the Fog of Love environment. We begin with establishing a baseline with MADDPG, followed by global affinities, and localized affinities. In the case of global affinities, we compare multiple scenarios such as agents having affinities towards a single action such as \textit{option 1}, or equal affinities towards all actions. For localized affinities, we compare agents with affinities towards single, multiple and cooperative virtues. All these combinations are in turn measured upto the MADDPG baseline. Using the metrics discussed in section \ref{fol}, we evaluate these agent performances.

\subsection{Global affinities}

We initially trained a vanilla MADDPG algorithm with two agents with no affinities. The success rate and goal errors are calculated across 24 simulations and then averaged. We then used the policy regularization with different combinations of action affinities for each player. For example, player 1 was assigned an affinity towards selecting option 1, and player towards option 3. 

\begin{table}[]
    \centering
    
    \begin{tabular}{|c|c|c|c|c|} \hline 
         \textbf{Player 1 affinity}&  \textbf{Player 2 affinity} & \textbf{Success rate ($\uparrow$)}&\textbf{Goal error ($\downarrow$)} &\textbf{Test score ($\uparrow$)}\\ \hline \hline 
         baseline&baseline& 0.39 & 77.73 & 29.69\\ \hline 
         equal & equal & 0.57 & 26.38 & 34.79 \\ \hline
 option 2& option 1& 0.36&48.56&38.5\\ \hline 
 option 1& option 2& 0.43&42.11&38.1\\ \hline 
 option 3& option 2& 0.40&48.84&40.1\\ \hline 
 option 2& option 3& 0.47&34.16&44.5\\ \hline 
 option 3& option 1& 0.49&34.34&\textbf{44.8}\\\hline 
 option 1& option 3& 0.40&48.42&39.6\\ \hline 
 option 1& option 1& \textbf{0.68}&\textbf{15.67}&15.1\\ \hline 
 option 2& option 2& \textbf{0.69}&\textbf{17.55}&23.7\\ \hline 
 option 3& option 3& \textbf{0.52}&\textbf{25.48} &31.7\\ \hline
 \end{tabular}    \caption{\textbf{Performance of Ab-RL with Global Affinities versus baseline agents:} Comparison between baseline and different combinations of action affinities. The regularization probabilities are $(0.6, 0.2, 0.2)$ for an agent favoring option 1. Players with affinities for the same action yield higher success rate and lower goal error. }
    \label{tab:global_affinity}
\end{table}

\begin{figure}
    \centering
    \begin{subfigure}[b]{0.45\linewidth}
        \includegraphics[width=\linewidth]{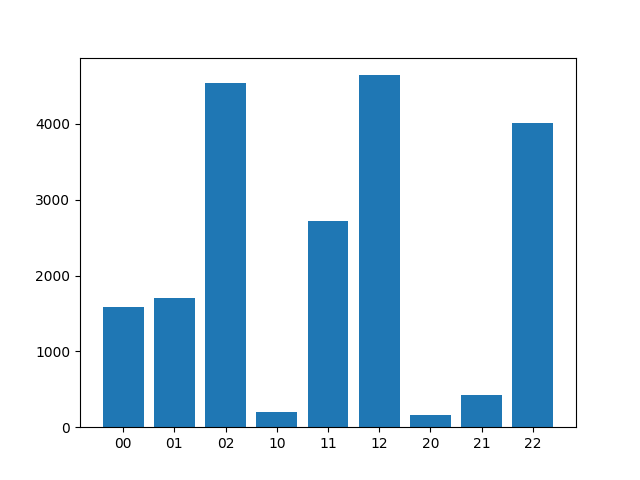}
        \caption{Baseline}
        \label{fig:baseline_act_dist}
    \end{subfigure}
    \begin{subfigure}[b]{0.45\linewidth}
        \includegraphics[width=\linewidth]{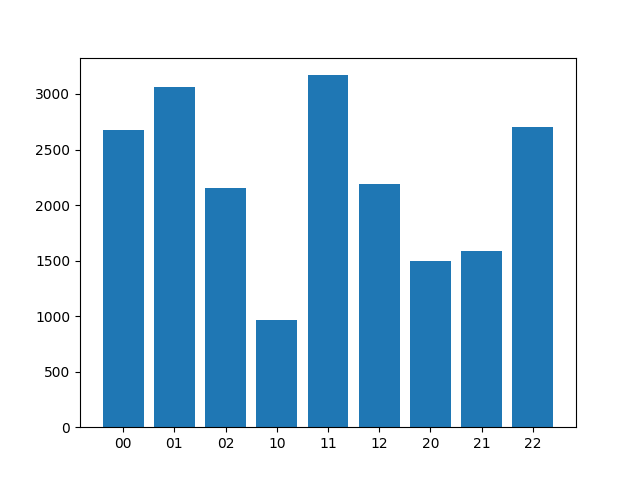}
        \caption{Both agents with equal affinities.}
        \label{fig:equal_affinity}
    \end{subfigure}
    \begin{subfigure}[b]{0.45\linewidth}
        \includegraphics[width=\linewidth]{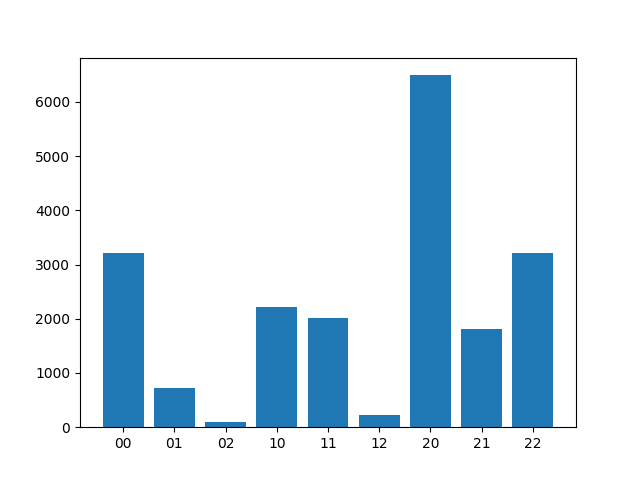}
        \caption{Option 3 and option 1 affinities.}
        \label{fig:different_affinity}
    \end{subfigure}
    \begin{subfigure}[b]{0.45\linewidth}
        \includegraphics[width=\linewidth]{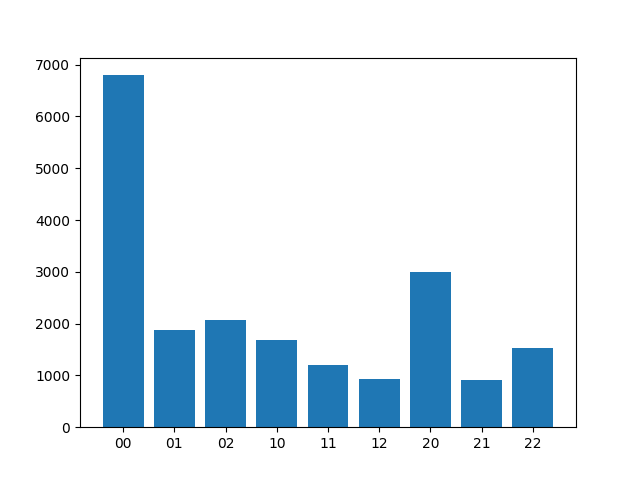}
        \caption{Both agents with option 1 affinity.}
        \label{fig:same_affinity}
    \end{subfigure}
    \caption{Action distribution ($0=\text{option 1}; 1=\text{option 2}; 2=\text{option 3}$) for baseline, different global affinities and same global affinities, respectively. The x-axis and y-axis represent the affinity combination and number of times this combination was recorded, respectively. Affinity-based MADDPG predisposes an agent to prefer certain actions while also optimizing the problem at hand. The figures show the respective action distributions across 200 test episodes, with each episode containing 100 \textit{scenes}.}
    \label{fig:action_distribution}
\end{figure}

In Table \ref{tab:global_affinity}, we present the performance of the agents by comparing their average success rates, goal errors and test scores. We can observe that the agents with global affinities have a consistently higher success rate and lower goal error. An interesting observation from Table \ref{tab:global_affinity} is the performance of agents when their action affinities are \textit{equal}, i.e., (0.33, 0.33, 0.33). It is evident that such agents perform better than the baseline across all metrics. Theoretically, agents with equal affinities, can be viewed as a typical regularization term which reduces overfitting and finds a balance between bias and variance. In Figure \ref{fig:equal_affinity}, it can be seen that the \textit{equal} affinities agents tend to distribute their actions more evenly in comparison to the baseline (Figure \ref{fig:baseline_act_dist}) and other types of affinities.

Next, when player 1 and player 2 have the same affinities, the performance is better than when they have different affinities. This is due to the design of Fog of Love, where the cooperative goal is defined such that when both players choose the same action, they score higher in \textit{satisfaction}. Thus they are likely to have a better success rate and a lower goal error, since they almost always accomplish their cooperative goals. While the agents with the same affinities excel at scoring in the metrics we defined, their test scores suffer due to their focus on the cooperative goal. We analysed this phenomenon further in Figures \ref{fig:different_affinity} and \ref{fig:same_affinity}, where we observe that when the \textit{same affinity} agents do not explore other action combinations as much as the \textit{different affinities} agents. Hence, the \textit{same affinities} agents do not improve on their test performance, which is the average reward obtained in 200 test episodes, while the \textit{different affinities} agents are more open to exploring options that yield higher reward.

\subsection{Localized affinities}
Here, we present the comparison between baseline and localized affinities in Table \ref{tab:localized}. For example if player 1 and player 2 have an affinity towards the virtue \textit{discipline}, it means that Algorithm \ref{alg:condition} has been used to conditionally check if the agent is actively excelling at \textit{discipline}, and if it does not, then a regularization loss is incurred. 

\begin{table}[]
    \centering
    \begin{tabular}{|p{1.7cm}|p{1.7cm}|>{\centering\arraybackslash}p{1.2cm}|>{\centering\arraybackslash}p{1.2cm}|>{\centering\arraybackslash}p{1.2cm}|>{\centering\arraybackslash}p{1.2cm}|} \hline 
         \textbf{Player 1 virtue}&  \textbf{Player 2 virtue} & \textbf{Success rate ($\uparrow$)} & \textbf{Affinity success rate ($\uparrow$)} &\textbf{ Goal error ($\downarrow$)}& \textbf{Test score ($\uparrow$)} \\ \hline\hline 
 baseline& baseline    & 0.39& - & 77.73 & 29.69\\ \hline
 discipline& discipline& 0.58 & 0.99 & 60.23 & \textbf{35.37}\\ \hline 
 gentleness& gentleness& \textbf{0.63} & 0.98 & \textbf{30.47} & 31.45\\ \hline 
 sensitivity& sincerity& 0.60 & 0.99 & 34.01 & 34.48\\ \hline 
 extraversion&curiosity& 0.61 & 0.99 & 30.72 & 35.34\\ \hline
    \end{tabular}
    \caption{\textbf{Performance of Ab-RL with Localized Affinities versus baseline agents:} Comparison of average success rate, affinity success rate and goal error, between baseline, action affinity and localized affinity. The virtues are \textit{discipline}, \textit{curiosity}, \textit{extraversion}, \textit{sensitivity}, \textit{gentleness}, and \textit{sincerity}. Localized-affinities with same virtues for both the players are averaged for 10 trials, and for different virtue affinities, they are averaged for 5 trials.}
    \label{tab:localized}
\end{table}

It is evident from Table \ref{tab:localized} that the agents with localized affinities have a higher success rate, lower goal error and better test scores than the baseline agents. This is regardless of whether the agents have the same affinities or not. When we calculate the success rate for a virtue corresponding to the agent's affinity, it results in a values close to 100\%, which is a guarantee that an agent would excel at the given virtue. However, this applies only to agents with a single virtue as an affinity. Therefore, we explored further by instilling the agent with multiple affinities.

\begin{table}[ht]
    \centering
    \begin{tabular}{|p{1.5cm}|p{1.5cm}|>{\centering\arraybackslash}p{1.2cm}|>{\centering\arraybackslash}p{1.6cm}|>{\centering\arraybackslash}p{1.6cm}|>{\centering\arraybackslash}p{1.2cm}|>{\centering\arraybackslash}p{1.2cm}|} \hline 
         \textbf{Player 1 virtues}&  \textbf{Player 2 virtues} & \textbf{Success rate ($\uparrow$)}& \textbf{Affinity success rate ($\uparrow$)}&  \textbf{Non-affinity success rate ($\uparrow$)}&\textbf{Goal error ($\downarrow$)}&\textbf{Test score ($\uparrow$)}\\ \hline \hline 
         baseline& 
    baseline& 0.39& - & - &77.73 & 29.69\\ \hline
 -& d,c,e,se,g,si& 0.59 & 0.86 & 0.64 &34.32 & \textbf{38.9}\\ \hline 
 d,c,e,se,g,si& -& 0.57& 0.98& 0.54 & 58.18 & 21.54\\ \hline 
 d,c,si& g,si& \textbf{0.63} & 0.99 & 0.71 &  \textbf{29.61} & 33.34\\ \hline 
 d,c & c,g,si& 0.58 &0.98 & 0.71 & 35.74 & 36.66\\ \hline
    \end{tabular}
    \caption{\textbf{Performance of Ab-RL with Multiple Localized Affinities:} Comparison of success rate, affinity success rate and goal error, between baseline and combinations of localized affinities. The virtues are \textit{discipline} (d), \textit{curiosity} (c), \textit{extraversion} (e), \textit{sensitivity}(se), \textit{gentleness} (g), and \textit{sincerity} (si). The values for localized affinities are averaged across 7 trials.}
    \label{tab:multi_aff}
\end{table}

In Table \ref{tab:multi_aff}, we show four different combinations of affinities, each averaged across seven trials. It can be observed that the success rates for the given affinities are still high, while being competent in the overall success rates and goal error values, thus, validating our result empirically. Note that for three out of the four combinations, the affinity success rates are almost 100\%, while for the case where Player 2 virtues are discipline (d), curiosity (c), extraversion (e), sensitivity (se), gentleness (g), and sincerity (si), the affinity success rate is 86\%. We refer to this as the \textit{all and no affinities} scenario, which is an extreme version where one player has no affinities, while the other player has to grapple with optimizing every virtue. This reduction in affinity score can happen because of the conflict between affinities, i.e., an action can have a positive and a negative effect on the affinities to be optimized. In addition, we can observe the test scores for the \textit{all and no affinities} scenario reflect the priorities of the agents, where agents might choose to prioritize affinities over reward, or vice versa. Also of importance is the cooperative goal, which we did not mention so far, because cooperative choices also result in higher rewards. Hence, this is a preliminary demonstration of the tension between cooperative and competitive goals, and the agents choosing to prioritize one over the other.

\subsection{Cooperative Goal}
Here, we demonstrate the performance of localized ab-RL on cooperative goals. We show the cooperative goal performance separately because this endeavor needed a different hyperparameter configuration compared to our previous experiments. Although experimentation with combinations of cooperative and competitive affinities remains theoretically feasible and implementable, we will defer this discussion in a future work. In Table \ref{tab:coop} we compare an agent trained to prefer the virtue \textit{satisfaction} over the competitive goals, and compare the same with our baseline performance. We can observe that the cooperative agents score significantly better at the metrics we defined to better capture an agent's performance in Fog of Love. While the cooperative agents score similar to baseline in the competitive goals, their score on \textit{satisfaction} is significantly better at 95\%. This means that 95\% of the time, the agents fulfill their cooperative goals. While it is evident that the average test episode scores are significantly lower or negative, this does not mean that the agents are playing Fog of Love poorly. Rather, when presented with a dilemma between choosing the same action as the other player, and maximizing reward, the agent prefers the former.

\begin{table}[ht]
    \centering
    \begin{tabular}{|p{5cm}|c|c|}
    \hline
        Metric & Baseline & Cooperative agent \\
        \hline
        \hline
Success rate & 0.39 & \textbf{0.652016} \\
Goal error ($\downarrow$) & 77.73 & \textbf{13.134167} \\
Average test score   & \textbf{29.69} & -8.02 \\
Satisfaction success rate &  0.08&     \textbf{0.95} \\
Discipline success rate &    0.51&     0.54 \\
Curiosity success rate &     0.53&     0.52 \\
Extraversion success rate &  0.53&     0.50 \\
Sensitivity success rate &   0.51&     0.54 \\
Gentleness success rate &    0.53&     0.51 \\
Sincerity success rate &     0.51&     0.55 \\
Overall competitive success rate & 0.52 &      \textbf{0.53} \\
        \hline

    \end{tabular}
    \caption{\textbf{Performance of agents with cooperative affinities \textit{(satisfaction)} versus baseline agents:} measured in terms of individual virtue success rate. The regularization strength ($\lambda$) was fixed at 1000. The cooperative agent records a higher satisfaction success rate, lower goal error and higher overall success rate.}
    \label{tab:coop}
\end{table}

\begin{figure}
\centering
    \begin{subfigure}[b]{0.45\linewidth}
        \includegraphics[width=\linewidth]{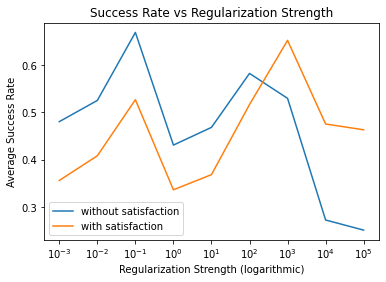}
        \caption{Success Rate}
        \label{fig:success_rate}
    \end{subfigure}
    \begin{subfigure}[b]{0.45\linewidth}
        \includegraphics[width=\linewidth]{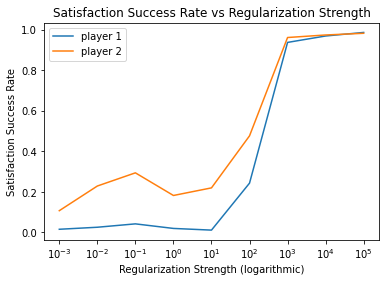}
        \caption{Satisfaction success rate}
        \label{fig:satisfaction_success}
    \end{subfigure}
    \begin{subfigure}[b]{0.45\linewidth}
        \includegraphics[width=\linewidth]{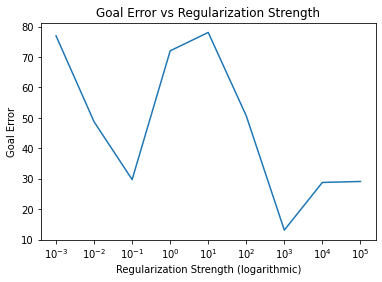}
        \caption{Goal Error}
        \label{fig:goal_error}
    \end{subfigure}
    \begin{subfigure}[b]{0.45\linewidth}
        \includegraphics[width=\linewidth]{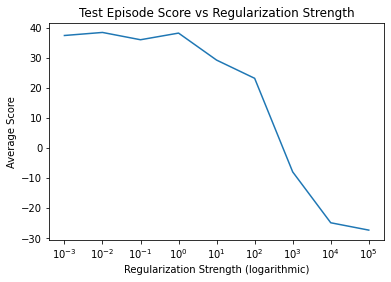}
        \caption{Test score}
        \label{fig:test_score}
    \end{subfigure}
    \caption{\textbf{Cooperative goal performance vs regularization strength:} The four figures illustrate the performance of agents with an affinity towards cooperative goals in a logarithmic scale. We present a) the overall success rate with and without \textit{satisfaction} in the calculation, b) average \textit{satisfaction} success rate for each agent, Player 1 and Player 2, c) Average goal error and d) average test score.}
    \label{fig:coop}
\end{figure}

The agents presented in the Table \ref{tab:coop}, require a regularization strength, $\lambda$, significantly higher than competitive agents (1000 vs. 5). This reveals the environment dynamics, wherein an agent is less likely to be cooperative and needs additional incentives. To explore how much incentive is required, we compared our metrics by varying $\lambda$ in Figure \ref{fig:coop}. The average success rate of the cooperative agents results in two peaks as we increase $\lambda$ logarithmically between $10^{-3}$ and $10^5$ (Figure \ref{fig:success_rate}) at $\lambda=10^{-1}$ and $\lambda=10^{3}$.  An explanation for two peaks in Figure \ref{fig:success_rate} is due to the effect of regularization due to lower and higher $\lambda$. Lower $\lambda$ motivates an agent to regularize the reward function, without much effect on the individual virtues or the cooperative goals. This effect was also observed in Table \ref{tab:global_affinity} where global affinities improved the success rate and goal error metrics, especially when the agents chose the same actions.

Now, if we remove \textit{satisfaction success rate} from our calculation, the second peak occurs at  $\lambda=10^{2}$. Also the overall trend for the average success rate, including and excluding \textit{satisfaction}, shows an overall increase and decrease, respectfully, demonstrating the effect of $\lambda$ until its value is $10^{4}$. After that, we should expect to see a decline, since the extremely high regularization penalty becomes an obstacle to learning other aspects of Fog of Love. In Figure \ref{fig:satisfaction_success}, we can validate the effect of $\lambda$ on the satisfaction success rate, which tends to 100\% as we increase it further.

The final figures \ref{fig:goal_error} and \ref{fig:test_score} show a steady decline as $\lambda$ is increased logarithmically. However, they mean very different things in these figures. For instance, the reduction in goal error (Figure \ref{fig:goal_error}) is due to the design of our version of Fog of Love. The cooperative goals are harder to achieve in comparison to the competitive goals. For example, in an episode, an agent would have to achieve \textit{discipline} $=4$ and \textit{sensitivity} $=-2$, but should always try to gain a \textit{satisfaction} $=30$, which is a lot more challenging. Hence, since a cooperative agent almost always fulfills 30 or more \textit{satisfaction}, the goal error is lower, while not achieving competitive goals does not result in a higher goal error. In other words, the cost of failing at competitive goals is lower than failing at the cooperative ones. Finally, the test episode score shows a steady decline because it measures only rewards achieved, and as we increase $\lambda$, it prioritizes the regularization term. 

\section{Discussion}\label{discussion} 

We began this article by discussing virtue ethics, an ancient theory of morality which saw a resurgence since the middle of the 20th century \cite{anscombe_modern_1958}. It posits that virtues are central to the endeavor of good moral character, rather than duties or consequences. In Section \ref{virtues}, we succinctly summarized MacIntyre's virtue theory that extends Aristotle's theory, by containing virtues within the context of social practices and being applicable across moral traditions. We posit that rather than evaluating AI algorithms with standard evaluation metrics, we could define additional metrics that examine the extent to which virtues are fulfilled. Engineering virtuous AI this way makes its role and teleology within a social practice coherent.

We then argued that ab-RL has been proposed as a paradigm to realize virtuous behavior in artificial agents, along with which we made a case for systemic role-playing games as framework to develop environments where an agent can be trained to be virtuous using moral dilemmas (as originally proposed in \cite{vishwanath_towards_2022}). We brought these three aspects, i.e., virtue ethics, RL, and systemic role-playing games together by engineering an RL environment based on the two player role-playing game called Fog of Love \cite{jacob_jaskov_fog_2017}, played by ab-RL agents.

The moral dilemmas in Fog of Love are choosing between selfish goals, such as fulfilling a player's character goals, and selfless goals, such as prioritizing the relationship with the other player. The original board game ends with both players achieving their \textit{destiny}, where they need to achieve certain trait goals and \textit{satisfaction} scores. In our implementation, we adopted the ``equal partners'' destiny, where both players must score 30 \textit{satisfaction} points. We then defined evaluation metrics such as \textit{success rate} and \textit{goal error} which measure how well it pursued their \textit{destiny}. Initially, the agents in the game are not informed of each other's individual objectives; however, as the game progresses, the decisions made by the other agent reveal these objectives. Consequently, to facilitate cooperation, the agents can make altruistic decisions that take into account the selfish objectives of their opponent. This is the systemic element of Fog of Love.

We develop a localized affinity-based RL \cite{vishwanath_localized_2025} algorithm on top of a multi-agent actor-critic algorithm known as MADDPG \cite{lowe2020multiagentactorcriticmixedcooperativecompetitive}. MADDPG employs distinct actor models for each agent while utilizing a shared critic, which acquires knowledge of the systemic characteristics of the environment through its comprehensive, higher-level perspective. We then outlined a novel algorithm (Algorithm \ref{alg:condition}) specific to Fog of Love, which computes the regularization term in Equation \ref{eqn:state_affinity} depending on the agent's state. Using this, we implemented virtuous behavior\footnote{\textit{virtuous}, in the AI sense, rather than virtues of humans. We stress that human and AI virtues are not the same, rather AI virtues are virtues in the functional sense. See \cite{wallach_moral_2010,powers_artificial_2017} for treatises on \textit{functional morality}.} by motivating an agent to prefer actions that fulfill the said virtue. We then compared the performance of baseline agents (without affinities) with agents instilled with global and localized affinities, using the metrics we defined.

Our results show that global affinities behave like the classic L1 and L2 regularization in machine learning, since there were visible performance improvements in goal achievement. In particular, when the regularization probabilities were equal $(0.33, 0.33, 0.33)$, the agent performed better across all metrics. Another interesting observation was when the agents were predisposed towards choosing the same actions, they scored even better success rates and goal errors. While choosing the same actions did not result in better test scores, the performance improvement on other metrics was significantly better. This highlights a trade-off between maximizing rewards versus choosing to be cooperative in the game. Other researchers \cite{maree_can_2022,vishwanath_exploring_2024} who have incorporated global affinities in their respective applications have demonstrated analogous outcomes. They also observed that applications based on ab-RL need to establish a balance between the extent of reward required for an agent to attain its objectives and the significance of other competencies within the application.

Next, we analyzed the localized affinity-based agents, which whether trained to excel at one virtue or more, almost always guaranteed higher success rates and lower goal errors. The trade-off, once again, was between the reward function and the virtues. Additionally, the virtues which the agents were predisposed to prefer are almost always resulted in a 100\% success rate for the said virtues. This is the most significant result of this article, since we confirm that it is possible to train an RL agent to excel at certain traits within a highly stochastic and complex environment, while simultaneously optimizing a reward function. Localized ab-RL has proven to be a powerful technique leveraged to excel at particular sub-goals without the need for sophisticated reward function design. Previous work \cite{vishwanath_localized_2025}, demonstrated the same, but in grid-world problems, while also highlighting its human-level interpretability. After all, to develop ethical agents, being transparent about how decisions were arrived at is critical.

Finally, the competition-cooperation dilemma was briefly explored. We highlighted some initial results between cooperative agents, where we achieved a cooperative success rate of 95\%, while also achieving comparable performance in competitive goals. An interesting future work in this domain is to empirically explore the interaction between a competitive and a cooperative. This is a problem adjacent to game theory about agents adopting an optimal strategy to obtain the best possible result in such encounters. Since we adopt multi-agent reinforcement learning for optimization, could ab-RL be proven to attain Pareto optimality? There has been encouraging results in the field of multi-agent RL which have worked on Pareto actor-critic models and deep RL models that converge to an optimal equilibrium \cite{christianos2023paretoactorcriticequilibriumselection,zhao_improvement_2023}. We could theoretically extend ab-RL in a similar way to achieve Pareto optimality, where policies of competing agents in Fog of Love could converge to an equilibrium, while also being in line with $\pi_0(A, S)$ (Equation \ref{eqn:state_affinity}).

\section{Conclusions and Directions for Future Work}\label{conc}

Affinity-based reinforcement learning is a novel approach to reinforcement learning which uses a superposition of prototypical behaviors as the basic building blocks of more complex behaviors with the aim of improving the interpretability of trained RL agents, i.e., the use of virtues within the ab-RL paradigm enables the model to be human-level interpretable. Rather than directly optimizing the reward function, ab-RL encourages the agent to learn policies that exhibit ``affinities'' towards certain desirable behaviors. These affinities are defined by policy regularization which is separate from the reward function.

Ab-RL is a promising framework for designing virtuous agents by shifting the focus beyond mere reward maximization. Instead of solely optimizing for a predefined reward function, ab-RL allows for the integration of virtues such as sensitivity, gentleness, discipline or curiosity as guiding principles for an agent's behavior. The use of prior probabilities of committing actions is analogous to emulating a virtuous role-model. The agent, in response to the diverse challenges presented by the environment, addresses these dilemmas by simultaneously optimizing the reward function and the regularization parameters. Virtuous behavior involves achieving a equilibrium between opposing extremes; full adherence to a role model is unnecessary, as individuals encounter distinct challenges, originate from varied backgrounds, and pursue virtuous behavior in their own unique manner.

In the context of multi-agent systems, the Nash equilibrium represents a condition in which no individual agent can improve its own outcome by unilaterally changing its strategy, provided the other agents' strategies remain constant; this focus on individual profit may result in suboptimal outcomes for the collective. Conversely, Pareto optimality defines a condition wherein no individual agent can be improved without making at least one other agent worse off. This concept more closely aligns with the objective of developing virtuous agents that prioritize the welfare of others and the common good, as Pareto optimality promotes: (1) collective flourishing, where agents that consider the well-being of other agents and the overall system, rather than solely focusing on maximizing their own rewards, (2) cooperation and collaboration, which encourage behaviors that yield better collective outcomes, and (3) ethical considerations that result in outcomes that are not only efficient but also fair and equitable for all agents involved.

While Nash equilibrium focuses on individual rationality, Pareto optimality prioritizes the collective flourishing. This aligns better with the ethical considerations and broader societal goals often associated with the engineering of artificial virtuous agents. Our future research will investigate whether we can demonstrate that ab-RL in fact converges in some sense to or near a Pareto optimal policy in multi-agent systems.

\backmatter




\section*{Declarations}

``Fog of Love''$^\circledR$ is a registered trademark of Floodgate Games, LLC. This research paper and its reinforcement learning environment are not affiliated with, endorsed by, or sponsored by Floodgate Games, LLC.

\bibliography{sn-bibliography}

@article{rodriguez-soto_instilling_2022,
	title = {Instilling moral value alignment by means of multi-objective reinforcement learning},
	volume = {24},
	issn = {1388-1957, 1572-8439},
	url = {https://link.springer.com/10.1007/s10676-022-09635-0},
	language = {en},
	number = {1},
	urldate = {2022-03-29},
	journal = {Ethics and Information Technology},
	author = {Rodriguez-Soto, Manel and Serramia, Marc and Lopez-Sanchez, Maite and Rodriguez-Aguilar, Juan Antonio},
	year = {2022},
	pages = {9},
}

@book{ross_oxford_1980,
	title = {{Aristotle}: {The} {Nicomachean} {Ethics} ({Revised} {Edition})},
	shorttitle = {Oxford {World}'s {Classics}},
 publisher = {Oxford {World}'s {Classics}},
	doi = {10.1093/actrade/9780199213610.book.1},
	language = {en},
    author={Ross, W. D.},
	editor = {Brown, Lesley},
	year = {1980},
    journal={Oxford {World}'s {Classics}},
    address={Oxford, U.K.}
}

@article{stenseke_artificial_2021,
	title = {Artificial virtuous agents: from theory to machine implementation},
	issn = {0951-5666, 1435-5655},
	shorttitle = {Artificial virtuous agents},
	url = {https://link.springer.com/10.1007/s00146-021-01325-7},
	doi = {10.1007/s00146-021-01325-7},
	language = {en},
	journal = {AI \& Society},
	author = {Stenseke, Jakob},
	year = {2021},
}

@article{stenseke_artificial_2022,
	title = {Artificial virtuous agents in a multi-agent tragedy of the commons},
	issn = {1435-5655},
	url = {https://doi.org/10.1007/s00146-022-01569-x},
	doi = {10.1007/s00146-022-01569-x},
	language = {en},
	urldate = {2022-11-28},
	journal = {AI \& Society},
	author = {Stenseke, Jakob},
	year = {2022},
}

@article{tolmeijer_implementations_2021,
	title = {Implementations in {Machine} {Ethics}: {A} {Survey}},
	volume = {53},
	issn = {0360-0300, 1557-7341},
	shorttitle = {Implementations in {Machine} {Ethics}},
	url = {https://dl.acm.org/doi/10.1145/3419633},
	doi = {10.1145/3419633},
	language = {en},
	number = {6},
	journal = {ACM Computing Surveys},
	author = {Tolmeijer, Suzanne and Kneer, Markus and Sarasua, Cristina and Christen, Markus and Bernstein, Abraham},
	year = {2021},
	pages = {1--38},
}

@book{wallach_moral_2010,
	address = {New York, NY},
	edition = {First issued as an Oxford University Press paperback},
	title = {Moral machines: teaching robots right from wrong},
	isbn = {978-0-19-973797-0},
	shorttitle = {Moral machines},
	language = {eng},
	publisher = {Oxford University Press},
	author = {Wallach, Wendell and Allen, Colin},
	year = {2010},
}

@inproceedings{nay_meaning_2017,
	address = {Hyannis Massachusetts},
	title = {Meaning without consequence: virtue ethics and inconsequential choices in games},
	isbn = {978-1-4503-5319-9},
	shorttitle = {Meaning without consequence},
	url = {https://dl.acm.org/doi/10.1145/3102071.3102073},
	doi = {10.1145/3102071.3102073},
	language = {en},
	booktitle = {Proceedings of the 12th {International} {Conference} on the {Foundations} of {Digital} {Games}},
	publisher = {ACM},
	author = {Nay, Jeff L. and Zagal, José P.},
	year = {2017},
	pages = {1--8},
}

@article{mnih2013playing,
  title={Playing atari with deep reinforcement learning},
  author={Mnih, Volodymyr},
  journal={arXiv preprint arXiv:1312.5602},
  year={2013}
}

@article{silver2018general,
  title={A general reinforcement learning algorithm that masters chess, shogi, and Go through self-play},
  author={Silver, David and Hubert, Thomas and Schrittwieser, Julian and Antonoglou, Ioannis and Lai, Matthew and Guez, Arthur and Lanctot, Marc and Sifre, Laurent and Kumaran, Dharshan and Graepel, Thore and others},
  journal={Science},
  volume={362},
  number={6419},
  pages={1140--1144},
  year={2018},
  publisher={American Association for the Advancement of Science}
}

@article{cervantes_artificial_2020,
	title = {Artificial {Moral} {Agents}: {A} {Survey} of the {Current} {Status}},
	volume = {26},
	issn = {1353-3452, 1471-5546},
	shorttitle = {Artificial {Moral} {Agents}},
	url = {http://link.springer.com/10.1007/s11948-019-00151-x},
	doi = {10.1007/s11948-019-00151-x},
	language = {en},
	number = {2},
	journal = {Science and Engineering Ethics},
	author = {Cervantes, José-Antonio and López, Sonia and Rodríguez, Luis-Felipe and Cervantes, Salvador and Cervantes, Francisco and Ramos, Félix},
	year = {2020},
	pages = {501--532},
}

@article{anscombe_modern_1958,
	title = {Modern {Moral} {Philosophy}},
	volume = {33},
	issn = {00318191, 1469817X},
	url = {http://www.jstor.org/stable/3749051},
	number = {124},
	urldate = {2024-04-17},
	journal = {Philosophy},
	author = {Anscombe, G. E. M.},
	year = {1958},
	note = {Publisher: Cambridge University Press},
	pages = {1--19},
}

@article{zhao_improvement_2023,
	title = {Improvement of {MADRL} {Equilibrium} {Based} on {Pareto} {Optimization}},
	volume = {66},
	copyright = {https://academic.oup.com/journals/pages/open\_access/funder\_policies/chorus/standard\_publication\_model},
	issn = {0010-4620, 1460-2067},
	url = {https://academic.oup.com/comjnl/article/66/7/1573/6567701},
	doi = {10.1093/comjnl/bxac027},
	language = {en},
	number = {7},
	urldate = {2025-01-31},
	journal = {The Computer Journal},
	author = {Zhao, Zhiruo and Cao, Lei and Chen, Xiliang and Lai, Jun and Zhang, Legui},
	month = jul,
	year = {2023},
	pages = {1573--1585},
}

@misc{christianos2023paretoactorcriticequilibriumselection,
      title={Pareto Actor-Critic for Equilibrium Selection in Multi-Agent Reinforcement Learning}, 
      author={Filippos Christianos and Georgios Papoudakis and Stefano V. Albrecht},
      year={2023},
      eprint={2209.14344},
      archivePrefix={arXiv},
      primaryClass={cs.LG},
      url={https://arxiv.org/abs/2209.14344}, 
}

@unpublished{berberich_virtuous_2018,
	title = {The {Virtuous} {Machine} - {Old} {Ethics} for {New} {Technology}?},
	url = {http://arxiv.org/abs/1806.10322},
	author = {Berberich, Nicolas and Diepold, Klaus},
	year = {2018},
	note = {arXiv: 1806.10322},
}

@book{crisp1997virtue,
  title={Virtue ethics},
  author={Crisp, Roger and Slote, Michael and Slote, Michael A},
  volume={10},
  year={1997},
  number={},
  address={Oxford, U.K.},
  publisher={Oxford University Press},
  journal={},
}

@article{maree_can_2022,
	title = {Can {Interpretable} {Reinforcement} {Learning} {Manage} {Prosperity} {Your} {Way}?},
	volume = {3},
	number = {2},
	journal = {AI},
	author = {Maree, Charl and Omlin, Christian W.},
	year = {2022},
	pages = {526--537},
}

@article{awad_moral_2018,
	title = {The {Moral} {Machine} experiment},
	volume = {563},
	issn = {0028-0836, 1476-4687},
	url = {https://www.nature.com/articles/s41586-018-0637-6},
	doi = {10.1038/s41586-018-0637-6},
	language = {en},
	number = {7729},
	urldate = {2025-02-05},
	journal = {Nature},
	author = {Awad, Edmond and Dsouza, Sohan and Kim, Richard and Schulz, Jonathan and Henrich, Joseph and Shariff, Azim and Bonnefon, Jean-François and Rahwan, Iyad},
	month = nov,
	year = {2018},
	pages = {59--64},
}

@inproceedings{peschl_moral_2022, 
    author = {Peschl, Markus and Zgonnikov, Arkady and Oliehoek, Frans A. and Siebert, Luciano C.}, 
    title = {MORAL: Aligning AI with Human Norms through Multi-Objective Reinforced Active Learning}, 
    year = {2022}, 
    isbn = {9781450392136}, 
    publisher = {International Foundation for Autonomous Agents and Multiagent Systems}, 
    address = {Richland, SC}, 
    booktitle = {Proceedings of the 21st International Conference on Autonomous Agents and Multiagent Systems}, 
    pages = {1038–1046},    
    numpages = {9}, 
    location = {Virtual Event, New Zealand}, 
    series = {AAMAS '22} 
}

@article{ozaki_finding_2024,
	title = {Finding middle grounds for incoherent horn expressions: the moral machine case},
	volume = {38},
	issn = {1387-2532, 1573-7454},
	shorttitle = {Finding middle grounds for incoherent horn expressions},
	url = {https://link.springer.com/10.1007/s10458-024-09681-6},
	doi = {10.1007/s10458-024-09681-6},
	language = {en},
	number = {2},
	urldate = {2025-02-05},
	journal = {Autonomous Agents and Multi-Agent Systems},
	author = {Ozaki, Ana and Rehman, Anum and Slavkovik, Marija},
	month = dec,
	year = {2024},
	pages = {50},
}

@article{tirumala_behavior_2022,
  author  = {Dhruva Tirumala and Alexandre Galashov and Hyeonwoo Noh and Leonard Hasenclever and Razvan Pascanu and Jonathan Schwarz and Guillaume Desjardins and Wojciech Marian Czarnecki and Arun Ahuja and Yee Whye Teh and Nicolas Heess},
  title   = {Behavior Priors for Efficient Reinforcement Learning},
  journal = {Journal of Machine Learning Research},
  year    = {2022},
  volume  = {23},
  number  = {221},
  pages   = {1--68},
  url     = {http://jmlr.org/papers/v23/20-1038.html}
}

@article{lang_utilitarian_2002,
	title = {Utilitarian {Desires}},
	volume = {5},
	issn = {13872532},
	url = {http://link.springer.com/10.1023/A:1015508524218},
	doi = {10.1023/A:1015508524218},
	number = {3},
	urldate = {2025-02-05},
	journal = {Autonomous Agents and Multi-Agent Systems},
	author = {Lang, Jérôme and Van Der Torre, Leendert and Weydert, Emil},
	year = {2002},
	pages = {329--363},
}

@misc{jacob_jaskov_fog_2017,
	title = {Fog of {Love}},
	url = {https://boardgamegeek.com/boardgame/175324/fog-of-love},
	language = {en-US},
	urldate = {2024-12-05},
	journal = {BoardGameGeek},
	author = {{Jacob Jaskov}},
	year = {2017},
}

@article{TH_sinnott-armstrong_consequentialism_2003,
	title = {Consequentialism},
	url = {https://plato.stanford.edu/entries/consequentialism/},
	urldate = {2022-11-17},
	author = {Sinnott-Armstrong, Walter},
	month = may,
	year = {2003},
}

@article{alexander_deontological_2021,
	title = {Deontological {Ethics}},
	url = {https://plato.stanford.edu/archives/win2021/entries/ethics-deontological/},
	urldate = {2022-04-05},
	journal = {The Stanford Encyclopedia of Philosophy},
	author = {Alexander, Larry and Moore, Michael},
	editor = {Zalta, Edward N.},
	year = {2021},
}

@article{formosa_papers_2016,
	title = {Papers, {Please} and the systemic approach to engaging ethical expertise in videogames},
	volume = {18},
	issn = {1388-1957, 1572-8439},
	url = {http://link.springer.com/10.1007/s10676-016-9407-z},
	doi = {10.1007/s10676-016-9407-z},
	language = {en},
	number = {3},
	journal = {Ethics and Information Technology},
	author = {Formosa, Paul and Ryan, Malcolm and Staines, Dan},
	year = {2016},
	pages = {211--225},
}

@article{hursthouse_virtue_2001,
	title = {On {Virtue} {Ethics}},
	url = {https://doi.org/10.1093/0199247994.001.0001},
	doi = {10.1093/0199247994.001.0001},
	author = {Hursthouse, Rosalind},
	month = sep,
	year = {2001},
}

@misc{cd_projekt_red_witcher_2015,
	title = {The {Witcher} 3: {Wild} {Hunt} - {Official} {Website}},
	shorttitle = {The {Witcher} 3},
	url = {https://www.thewitcher.com/en},
	urldate = {2022-04-11},
	author = {CD Projekt Red},
	month = may,
	year = {2015},
}

@misc{jonathan_gilmour_dead_2014,
	title = {Dead of {Winter}: {A} {Crossroads} {Game}},
	copyright = {Plaid Hat Games},
	url = {https://boardgamegeek.com/boardgame/150376/dead-of-winter-a-crossroads-game},
	urldate = {2024-12-05},
	author = {{Jonathan Gilmour} and {Isaac Vega}},
	year = {2014},
}

@misc{11_bit_studios_this_2014,
	title = {This {War} of {Mine}},
	url = {https://www.thiswarofmine.com/},
	language = {en-US},
	urldate = {2024-12-05},
	author = {{11 bit studios}},
	year = {2014},
	note = {This War Of Mine provides an experience of war seen from an entirely new angle. For the very first time you do not play as an elite soldier, rather a group of civilians trying to survive in a besieged city.},
}

@misc{zaum_disco_2019,
	title = {Disco {Elysium}},
	url = {https://discowp.zaumstudio.com},
	language = {en},
	urldate = {2024-12-05},
	author = {{ZA/UM}},
	year = {2019},
}

@incollection{powers_artificial_2017,
	address = {Cham},
	title = {Artificial {Moral} {Cognition}: {Moral} {Functionalism} and {Autonomous} {Moral} {Agency}},
	volume = {128},
	isbn = {978-3-319-61042-9 978-3-319-61043-6},
	shorttitle = {Artificial {Moral} {Cognition}},
	url = {http://link.springer.com/10.1007/978-3-319-61043-6\_7},
	language = {en},
	booktitle = {Philosophy and {Computing}},
	publisher = {Springer International Publishing},
	author = {Howard, Don and Muntean, Ioan},
	editor = {Powers, Thomas M.},
	year = {2017},
	doi = {10.1007/978-3-319-61043-6_7},
	pages = {121--159},
}

@inproceedings{govindarajulu_toward_2019,
	address = {Honolulu HI USA},
	title = {Toward the {Engineering} of {Virtuous} {Machines}},
	isbn = {978-1-4503-6324-2},
	url = {https://dl.acm.org/doi/10.1145/3306618.3314256},
	doi = {10.1145/3306618.3314256},
	language = {en},
	urldate = {2022-03-02},
	booktitle = {Proceedings of the 2019 {AAAI}/{ACM} {Conference} on {AI}, {Ethics}, and {Society}},
	publisher = {ACM},
	author = {Govindarajulu, Naveen Sundar and Bringsjord, Selmer and Ghosh, Rikhiya and Sarathy, Vasanth},
	year = {2019},
	pages = {29--35},
}

@article{moor_nature_2006,
	title = {The {Nature}, {Importance}, and {Difficulty} of {Machine} {Ethics}},
	volume = {21},
	issn = {1541-1672},
	url = {http://ieeexplore.ieee.org/document/1667948/},
	doi = {10.1109/MIS.2006.80},
	language = {en},
	number = {4},
	journal = {IEEE Intelligent Systems},
	author = {Moor, J.H.},
	year = {2006},
	pages = {18--21},
}

@inproceedings{ng1999policy,
  title={Policy invariance under reward transformations: Theory and application to reward shaping},
  author={Ng, Andrew Y and Harada, Daishi and Russell, Stuart},
  booktitle={Icml},
  volume={99},
  pages={278--287},
  year={1999},
  organization={Citeseer}
}

@inproceedings{achiam2017constrained,
  title={Constrained policy optimization},
  author={Achiam, Joshua and Held, David and Tamar, Aviv and Abbeel, Pieter},
  booktitle={International conference on machine learning},
  pages={22--31},
  year={2017},
  organization={PMLR}
}

@article{wirth2017survey,
  title={A survey of preference-based reinforcement learning methods},
  author={Wirth, Christian and Akrour, Riad and Neumann, Gerhard and F{\"u}rnkranz, Johannes and others},
  journal={Journal of Machine Learning Research},
  volume={18},
  number={136},
  pages={1--46},
  year={2017},
  publisher={Journal of Machine Learning Research/Massachusetts Institute of Technology~…}
}

@article{garcia2015comprehensive,
  title={A comprehensive survey on safe reinforcement learning},
  author={Garc{\i}a, Javier and Fern{\'a}ndez, Fernando},
  journal={Journal of Machine Learning Research},
  volume={16},
  number={1},
  pages={1437--1480},
  year={2015}
}

@article{vishwanath_towards_2022,
	title = {Towards artificial virtuous agents: games, dilemmas and machine learning},
	issn = {2730-5953, 2730-5961},
	shorttitle = {Towards artificial virtuous agents},
	language = {en},
	urldate = {2023-01-24},
	journal = {AI and Ethics},
	author = {Vishwanath, Ajay and Bøhn, Einar Duenger and Granmo, Ole-Christoffer and Maree, Charl and Omlin, Christian},
	year = {2022},
	pages = {s43681--022--00251--8},
}

@article{persiani2022policy,
  title={Policy regularization for legible behavior},
  author={Persiani, Michele and Hellstr{\"o}m, Thomas},
  journal={Neural Computing and Applications},
  pages={1--10},
  year={2022},
  publisher={Springer}
}

@book{macintyre_after_2007,
	title = {After virtue: {A} {Study} in {Moral} {Theory}},
	shorttitle = {After virtue},
	language = {en},
	author = {MacIntyre, Alasdair C.},
	year = {2007},
}

@article{maree_reinforcement_2022,

title = {Reinforcement {Learning} {Your} {Way}: {Agent} {Characterization} through {Policy} {Regularization}},
	volume = {3},
	issn = {2673-2688},
	shorttitle = {Reinforcement {Learning} {Your} {Way}},
	number = {2},
	journal = {AI},
	author = {Maree, Charl and Omlin, Christian},
	year = {2022},
	pages = {250--259},
}

@inproceedings{vishwanath_exploring_2024,
	address = {Singapore},
	title = {Exploring {Affinity}-{Based} {Reinforcement} {Learning} for {Designing} {Artificial} {Virtuous} {Agents} in {Stochastic} {Environments}},
	isbn = {978-981-9998-36-4},
	booktitle = {Frontiers of {Artificial} {Intelligence}, {Ethics}, and {Multidisciplinary} {Applications}},
	publisher = {Springer Nature Singapore},
	author = {Vishwanath, Ajay and Omlin, Christian},
	editor = {Farmanbar, Mina and Tzamtzi, Maria and Verma, Ajit Kumar and Chakravorty, Antorweep},
	year = {2024},
	pages = {25--38},
}

@inproceedings{reward_hacking,
author = {Skalse, Joar and Howe, Nikolaus H. R. and Krasheninnikov, Dmitrii and Krueger, David},
title = {Defining and characterizing reward hacking},
year = {2024},
isbn = {9781713871088},
publisher = {Curran Associates Inc.},
address = {Red Hook, NY, USA},
booktitle = {Proceedings of the 36th International Conference on Neural Information Processing Systems},
articleno = {687},
numpages = {12},
location = {New Orleans, LA, USA},
series = {NIPS '22}
}

@inproceedings{ran2023policy,
  title={Policy regularization with dataset constraint for offline reinforcement learning},
  author={Ran, Yuhang and Li, Yi-Chen and Zhang, Fuxiang and Zhang, Zongzhang and Yu, Yang},
  booktitle={International Conference on Machine Learning},
  pages={28701--28717},
  year={2023},
  organization={PMLR}
}

@article{DENNIS20161,
title = {Formal verification of ethical choices in autonomous systems},
journal = {Robotics and Autonomous Systems},
volume = {77},
pages = {1-14},
year = {2016},
issn = {0921-8890},
doi = {https://doi.org/10.1016/j.robot.2015.11.012},
url = {https://www.sciencedirect.com/science/article/pii/S0921889015003000},
author = {Louise Dennis and Michael Fisher and Marija Slavkovik and Matt Webster}
}

@article{Anderson_Anderson_2007, 
title={Machine Ethics: Creating an Ethical Intelligent Agent}, 
volume={28}, url={https://ojs.aaai.org/index.php/aimagazine/article/view/2065}, DOI={10.1609/aimag.v28i4.2065}, abstractNote={The newly emerging field of machine ethics (Anderson and Anderson 2006) is concerned with adding an ethical dimension to machines. Unlike computer ethics -- which has traditionally focused on ethical issues surrounding humans’ use of machines -- machine ethics is concerned with ensuring that the behavior of machines toward human users, and perhaps other machines as well, is ethically acceptable. In this article we discuss the importance of machine ethics, the need for machines that represent ethical principles explicitly, and the challenges facing those working on machine ethics. We also give an example of current research in the field that shows that it is possible, at least in a limited domain, for a machine to abstract an ethical principle from examples of correct ethical judgments and use that principle to guide its own behavior.}, 
number={4}, 
journal={AI Magazine}, 
author={Anderson, Michael and Anderson, Susan Leigh}, year={2007}, 
month={Dec.}, 
pages={15} 
}

@incollection{guarda_machine_2024,
	address = {Cham},
	title = {Machine {Ethics} and the {Architecture} of {Virtue}},
	volume = {1936},
	isbn = {978-3-031-48854-2 978-3-031-48855-9},
	url = {https://link.springer.com/10.1007/978-3-031-48855-9\_29},
	language = {en},
	urldate = {2024-10-26},
	booktitle = {Advanced {Research} in {Technologies}, {Information}, {Innovation} and {Sustainability}},
	publisher = {Springer Nature Switzerland},
	author = {Ribeiro, Beatriz A. and Da Silva, Maria Braz},
	editor = {Guarda, Teresa and Portela, Filipe and Diaz-Nafria, Jose Maria},
	year = {2024},
	doi = {10.1007/978-3-031-48855-9_29},
	note = {Series Title: Communications in Computer and Information Science},
	pages = {384--401},
}

@misc{lowe2020multiagentactorcriticmixedcooperativecompetitive,
      title={Multi-Agent Actor-Critic for Mixed Cooperative-Competitive Environments}, 
      author={Ryan Lowe and Yi Wu and Aviv Tamar and Jean Harb and Pieter Abbeel and Igor Mordatch},
      year={2020},
      eprint={1706.02275},
      archivePrefix={arXiv},
      primaryClass={cs.LG},
      url={https://arxiv.org/abs/1706.02275}, 
}

@inproceedings{vishwanath_localized_2025,
	address = {Cham},
	title = {Localized {Affinity}-{Based} {Reinforcement} {Learning} for {Interpretable} {State}-{Specific} {Decision}-{Making}},
	isbn = {978-3-031-77915-2},
	booktitle = {Artificial {Intelligence} {XLI}},
	publisher = {Springer Nature Switzerland},
	author = {Vishwanath, Ajay and Omlin, Christian},
	editor = {Bramer, Max and Stahl, Frederic},
	year = {2025},
	pages = {221--234},
}

@article{adams_survey_2022,
	title = {A survey of inverse reinforcement learning},
	volume = {55},
	issn = {1573-7462},
	url = {https://doi.org/10.1007/s10462-021-10108-x},
	doi = {10.1007/s10462-021-10108-x},
	number = {6},
	journal = {Artificial Intelligence Review},
	author = {Adams, Stephen and Cody, Tyler and Beling, Peter A.},
	month = aug,
	year = {2022},
	pages = {4307--4346},
}

@article{zhong_computational_2025,
    title = {Computational {Machine} {Ethics}: {A} {Survey}},
    volume = {82},
    issn = {1076-9757},
    shorttitle = {Computational {Machine} {Ethics}},
    url = {https://www.jair.org/index.php/jair/article/view/16836},
    doi = {10.1613/jair.1.16836},
    urldate = {2025-03-20},
    journal = {Journal of Artificial Intelligence Research},
    author = {Zhong, Tammy and Song, Yang and Limarga, Raynaldio and Pagnucco, Maurice},
    month = mar,
    year = {2025},
    pages = {1581--1628},
}



\newpage

\begin{appendices}

\section{Observation Space}\label{state-space}

In this section, we outline the structure of the state space. The virtues are encoded as numbers between 1 and 6, and \textit{satisfaction} is represented by 7. The virtues are as follows:
\begin{enumerate}
    \item discipline
    \item curiosity
    \item extraversion
    \item sensitivity
    \item gentleness
    \item sincerity
\end{enumerate}

In Table \ref{tab:statespace}, we summarize the observation space of an agent in Fog of Love.

\begin{longtable}{>{\ttfamily}llp{4.2cm}ll}
\toprule
\multicolumn{4}{c}{Agent Observation Space Structure} \\
\midrule
Agent & Key Prefix & Description & Range & Type \\
\midrule
\endfirsthead
\multicolumn{4}{c}{\textit{Continued from previous page}} \\
\midrule
Agent & Key Prefix & Description & Range & Type \\
\midrule
\endhead
\midrule
\multicolumn{4}{c}{\textit{Continued on next page}} \\
\endfoot
\bottomrule
\endlastfoot

player 1 & goal\_* & Goal values (1-7) & [-50,50] & int32 \\
player 1 & player\_1\_virtue\_* & Player 1 virtue values (1-7) & [-50,50] & int32 \\
player 1 & player\_2\_virtue\_* & Player 2 virtue values (1-7) & [-50,50] & int32 \\
player 1 & option1\_virtue\_* & Option 1 values (1-7) & [-1,1] & int32 \\
player 1 & option2\_virtue\_* & Option 2 values (1-7) & [-1,1] & int32 \\
player 1 & option3\_virtue\_* & Option 3 values (1-7) & [-1,1] & int32 \\
player 1 & match & If both players choose the same option & [-1,1] & int32 \\
player 1 & no\_match & If both players choose different options & [-1,1] & int32 \\
\addlinespace
player 2 & goal\_* & Goal values (1-7) & [-50,50] & int32 \\
player 2 & player\_1\_virtue\_* & Player 1 virtue values (1-7) & [-50,50] & int32 \\
player 2 & player\_2\_virtue\_* & Player 2 virtue values (1-7) & [-50,50] & int32 \\
player 2 & option1\_virtue\_* & Option 1 values (1-7) & [-1,1] & int32 \\
player 2 & option2\_virtue\_* & Option 2 values (1-7) & [-1,1] & int32 \\
player 2 & option3\_virtue\_* & Option 3 values (1-7) & [-1,1] & int32 \\
player 2 & match & If both players choose the same option & [-1,1] & int32 \\
player 2 & no\_match & If both players choose different options & [-1,1] & int32 \\
\caption{Table representing the observation space of agents \textit{player 1} and \textit{player 2}.}
\label{tab:statespace}
\end{longtable}

The first row represents the trait and satisfaction goals of \textit{player 1}, that are only known to \textit{player 1}. The subsequent two rows represent the current value of virtues achieved upon playing the game. The next three rows represent the current options presented to \textit{player 1}, with each option containing the six virtues and \textit{satisfaction} values. The last two observations for \textit{player 1} are based on the \textit{match condition}, i.e., the amount of \textit{satisfaction} added or penalized depending on whether \textit{player 1} and \textit{player 2} choose the same option. In total, there are 44 observation variables for each player. The current (and goal) virtue and \textit{satisfaction} values range between -50 and +50, while the option and match condition values range between -1 and 1. 

\section{Hyperparameters}

Below we enlist the hyperparameter configuration utilized to train the MADDPG algorithm for all agents. The remaining hyperparameters are default to the AgileRL library default configurations\footnote{\href{https://docs.agilerl.com/en/latest/api/algorithms/maddpg.html}{https://docs.agilerl.com/en/latest/api/algorithms/maddpg.html}}. 

\begin{table}[h]
    \centering
    \begin{tabular}{|p{4cm}|c|}
    \hline
    Hyperparameter & Value \\
    \hline
    Hidden layer size  & [1024, 1024, 1024] \\
    Actor learning rate & $1e^{-5}$ \\
    Critic learning rate & $1e^{-2}$ \\
    Epochs & 10 \\
    Learning steps & 5 \\
    Batch size & 64 \\
    Discount factor, $\gamma$ & 0.95 \\
    Soft update of target network parameters, $\tau$  & 0.01 \\
    \hline
    \end{tabular}
    \caption{Hyperparameters used to train MADDPG.}
    \label{tab:hyperparameters}
\end{table}

\end{appendices}

\end{document}